\documentclass[11pt, a4paper]{custom_class}

\usepackage{custom_style}
\usepackage[utf8]{inputenc} 
\usepackage[T1]{fontenc}    
\usepackage{hyperref}       
\usepackage{url}            
\usepackage{booktabs}       
\usepackage{amsfonts}       
\usepackage{nicefrac}       
\usepackage{microtype}      
\usepackage{cleveref}       
\usepackage{lipsum}         
\usepackage{graphicx}
\usepackage{natbib}
\usepackage{doi}
\usepackage{natbib}
\newenvironment{displaytext}[1][2em]
  {\par\medskip               
   \begingroup                
   \setlength{\leftskip}{#1}
   \noindent\ignorespaces     
  }%
  {\par                       
   \endgroup\medskip}         

\title{The Vector Grounding Problem}

\author{
Dimitri Coelho Mollo \\
Umeå University \\
\texttt{dimitri.mollo@umu.se} \\
\And
Raphaël Millière \\
University of Oxford \\
\texttt{raphael.milliere@philosophy.ox.ac.uk} \\
}

\date{\textcolor{red}{Forthcoming in \textit{Philosophy and the Mind Sciences}}}

\renewcommand{\shorttitle}{The Vector Grounding Problem}

\begin{document}
\maketitle

\begin{abstract}
Large language models (LLMs) produce seemingly meaningful outputs, yet they are trained on text alone without direct interaction with the world. This leads to a modern variant of the classical symbol grounding problem in AI: can LLMs' internal states and outputs be about extra-linguistic reality, independently of the meaning human interpreters project onto them?  We argue that they can. We first distinguish referential grounding---the connection between a representation and its worldly referent---from other forms of grounding and argue it is the only kind essential to solving the problem. We contend that referential grounding is achieved when a system's internal states satisfy two conditions derived from teleosemantic theories of representation: (1) they stand in appropriate causal-informational relations to the world, and (2) they have a history of selection that has endowed them with the function of carrying this information. We argue that LLMs can meet both conditions, even without multimodality or embodiment.
\end{abstract}

\section{Introduction} \label{intro}

The remarkable performance of Large Language Models (LLMs) in sophisticated linguistic and cognitive tasks---once considered reliable indicators of human-like intelligence---has rekindled philosophical discussions on the nature of linguistic competence.\footnote{See \citet{Milliere2024} for a general introduction to language models, and \citet{Milliereforthcoming} for a discussion of their relevance to debates about linguistic competence}
Despite being trained exclusively on text data without direct interaction with the physical world, these models exhibit the ability to generate coherent and contextually relevant paragraphs, answer questions, and solve commonsense reasoning tasks in a wide range of domains. This ability raises intriguing questions about the relationship between language, meaning, and reference.\footnote{For recent discussion, see \citet{Butlin2021, cappelenMakingAIIntelligible2021, Pavlick_2022, chalmersDoesThoughtRequire2023, mandelkernLanguageModelsWords2024, Lederman2024, miracchititusDoesChatGPTHave2024, grindrodLargeLanguageModels2024, borgLLMsTuringTests, peppReferenceIntentionsLarge}.} 

In this paper, we revisit the classical `Symbol Grounding Problem' \citep{Harnad1990}, originally formulated for symbolic AI systems, in light of contemporary LLMs. The core issue can be summarized as follows: if AI systems process linguistic inputs without direct interaction with the world, can their internal representations and outputs possess any meaning beyond the interpretations that we, as intelligent beings embedded in the world, project onto them? We use the term \textit{intrinsic meaning} to capture this notion of interpretation-independent meaning, distinguishing it from meaning that is imposed from outside the system itself, which is thus extrinsic.\footnote{This follows \citet{Harnad1990}'s terminology. Intrinsic meaning, as we understand it here, is fully compatible with externalist views about how states and outputs come to acquire interpretation-independent meaning and reference. The intrinsic/extrinsic distinction cuts across the internalism/externalism one.} 
We argue that a similar conundrum arises for LLMs, which we call the \textit{Vector Grounding Problem}. Since these systems compute over continuous vectors rather than discrete symbols, the problem takes a somewhat different form. After outlining this new version of the classical problem, we suggest that there are strong reasons to believe that a basic form of grounding is possible in current LLMs. 

Specifically, we contend that the learning process of LLMs endows some of their internal states with functions that go beyond mere linguistic prediction. Although LLMs learn from text data, they undergo selection processes that favour internal states tracking features of the world whenever extra-linguistic criteria influence the training objective. This can occur directly by fine-tuning LLMs on objectives that reflect extra-linguistic norms---such as factuality---or indirectly via the implicit extra-linguistic success conditions of latent tasks embedded within the training data or provided in context. When internal states selected to carry information about specific worldly conditions causally influence the generation of linguistic outputs, those outputs become grounded in the world in a way that makes them meaningful independently of human interpretation. The crucial question is thus not whether outputs can be interpreted as meaningful by humans (they obviously can), but whether the outputs themselves stand in the right kinds of relations to the world such that they would be meaningful even in the absence of external interpretation.

Importantly, this paper does not address whether LLMs and related models \textit{understand} language (or anything at all), engage in linguistic acts, or possess cognitive or mental properties.\footnote{See \citet{goldsteinDoesChatGPTHave2024} and \citet{chalmersPropositionalInterpretabilityArtificial2025} for discussions that also draw on theories of content but consider whether LLMs can be ascribed propositional attitudes, and \citet{stoljarWhyChatGPTDoesnt2025} for a discussion of whether LLMs can think.} We do not claim that these models `know' what they mean, or even what meaning is, nor that they have any conception of truth or falsity. Our sole contention is that the problem of grounding internal states and outputs in the world is solvable in principle, and in some cases arguably already in practice, for LLMs and related models. While having grounded internal states may be a crucial step toward understanding, agency, knowledge, and various other cognitive and mental capacities, it is unlikely that such states are, by themselves, sufficient for those capacities. This clarification is especially important, as the current debate occasionally blurs these distinctions---a conflation we strive to avoid in this paper. 

We proceed as follows. First, we introduce the Symbol Grounding Problem in the context of classical AI (\autoref{classicalconnectionist}). Next, we present the Vector Grounding Problem as it applies to LLMs and related models (\autoref{largepretrained}). We then differentiate between various notions of grounding, showing that only one of them---\textit{referential grounding}---is essential to the Vector Grounding Problem (\autoref{fivenotions}). Drawing on philosophical theories of representational content, we articulate a set of conditions that a system must satisfy to achieve referential grounding (\autoref{theoriesrepresentation}). We argue that at least some LLMs and related multimodal models satisfy these requirements, and therefore have internal representations and outputs with intrinsic meaning (\autoref{groundinglargeAI}). Finally, we discuss some surprising implications of our argument (\autoref{sec:implications}).

\section{The Classical Symbol Grounding Problem} \label{classicalconnectionist}

Research in artificial intelligence has historically been shaped by two distinct paradigms: the classical (or symbolic) approach and the connectionist approach. The classical paradigm focuses on systems that represent and manipulate information through discrete symbols according to explicit rules. This symbolic manipulation is purely syntactic; it depends solely on the formal properties of symbols and the rules that govern their combination into complex structures \citep{simonHumanProblemSolving1971}. By contrast, the connectionist paradigm uses artificial neural networks that process information through massively parallel and distributed computation across interconnected nodes. Rather than following predetermined symbolic rules, these systems induce patterns directly from data through learning algorithms \citep{mcclellandParallelDistributedProcessing1986}.

The Symbol Grounding Problem, as identified by \citet{Harnad1990}, asks how the meaning of symbols in a formal symbol system can be intrinsic to the system itself, rather than merely parasitic on the meanings in the minds of external interpreters.
More precisely:

\begin{displaytext}
\textbf{Symbol Grounding Problem.} How can symbol tokens, manipulated solely on the basis of their shapes according to syntactic rules, acquire meaning and refer to entities and properties in the real world---rather than merely relating  to other meaningless symbol tokens---without depending on the semantic interpretations provided by external meaning-bearing systems?\footnote{It is common to take Harnad's Symbol Grounding Problem as an elaboration of an earlier, highly influential criticism of AI: \citet{Searle1980}'s Chinese Room thought experiment. \citet{Harnad1990} himself followed this line.} 
\end{displaytext}

\citet{Harnad1990} provides a compelling intuition pump to motivate
the Symbol Grounding Problem in classical symbolic AI. Consider a system---biological or artificial---with no prior knowledge of any natural language. Suppose we attempt to teach this system Chinese (or any other language), but provide it with only a monolingual resource: a
comprehensive Chinese-Chinese dictionary. Can the system
learn Chinese from this dictionary alone? The thought experiment invites a negative answer. The system can look up strings of symbols and find other strings that purportedly define them---words paired with their definitions. However, these symbols remain mere meaningless shapes followed by other shapes. Without any connection to the external world, it seems impossible for the system to grasp the actual meaning of these words. The system is trapped in a symbolic `merry-go-round', endlessly cycling through symbols that refer only to other symbols, never breaking free to establish connections with real-world referents. Without access to anything beyond symbols, how can the system ever ground these symbols in the real world?

According to Harnad, symbolic AI systems find themselves in this predicament. Limited to manipulating strings of symbols and their syntactic relationships, these systems lack any mechanism for establishing connections between symbols and their real-world referents. Consequently, the symbols possess no intrinsic meaning. While humans---who do have representations acquired through direct engagement with the world---can interpret these symbolic outputs as meaningful, this meaning remains extrinsic, entirely dependent on human interpretation. Remove the human interpreter, and the symbols become meaningless shapes. In essence, the argument shows that symbolic AI systems manipulate ungrounded symbols: symbols disconnected from the world they purport to represent. These systems thus lack intrinsic semantic content. Classical AI systems, in other words, are fundamentally limited by the Symbol Grounding Problem.

\section{The Vector Grounding Problem} \label{largepretrained}

Connectionist models have come a long way since the 1980s, with deep neural networks (DNNs) emerging as powerful tools for various tasks \citep{lecunDeepLearning2015}. Recent research has converged on training deep neural networks on vast amounts of data to perform a broad range of tasks they were not explicitly designed for. Such DNNs are often called `foundation models' \citep{bommasaniOpportunitiesRisksFoundation2022}, though we will use the more descriptive label `generative pre-trained models'. Most of these models are built with a neural network architecture called the `Transformer', introduced by \citet{Vaswani2017}, which uses a mechanism called `attention' to capture global dependencies between sequential elements. This architecture has proved particularly effective for linguistic and visual data.

LLMs are generative pre-trained models trained on text data. State-of-the-art LLMs have over \(10^{11}\) parameters and are trained on massive corpora spanning
trillions of words, effectively ingesting a substantial fraction of all text available on the internet. After training, LLMs
can generate well-written and coherent paragraphs about virtually any
topic. Furthermore, given explicit instructions in natural language in
the input sequence, or `prompts', they can perform a broad range of
tasks, including: summarising long-form content, translating between
multiple languages, answering complex questions, engaging in commonsense
reasoning, solving logic and maths problems, generating code, or explaining jokes \citep{srivastava2023beyond}.

LLMs process text into a sequence of \textit{tokens}: units that may map onto whole words or word parts. Each token is encoded as a high-dimensional vector called an \textit{embedding}. The model then estimates a probability distribution over possible next tokens based on the sequence of token embeddings. During training, next-token predictions are compared against the actual next token in the training data, and model parameters are adjusted to increase the probability of correct tokens while suppressing alternatives. Though LLMs do not treat words as the main units of processing, the combination of tokens can form an indefinite number of possible words, and we can view the vectors for such composing tokens as jointly forming a word embedding.  This training process (confusingly known as pre-training) encourages the model to produce context-sensitive embeddings,  with semantically similar words having vectors that are close together in the high-dimensional space.

There is a deep connection between the way in which LLMs learn
embeddings from linguistic corpora, and the old idea
from structural linguistics according to which word meaning can be
inferred from the contexts in which they appear, known as the
\textit{distributional hypothesis} \citep{harrisDistributionalStructure1954, firthPapersLinguistics193419511957}. In LLMs,
semantic similarity is determined based on distribution across contexts:
words that appear in similar contexts are mapped to nearby vectors.
Although individual dimensions of the vector space are not typically interpretable, the geometric relationships between vectors encode statistical regularities that reflect latent semantic structure. Consequently, embeddings are not purely conventional, unlike symbols in classical AI systems, as their relational structure is determined by statistical patterns in linguistic data.\footnote{Vision-Language Models (VLMs) extend this approach by encoding both textual and visual inputs within a unified vector space. We examine the implications of such multimodal architectures for the grounding problem in \autoref{multimodal-embodied}.} 

Nevertheless, LLMs seem to face a challenge analogous to that of traditional symbolic AI systems regarding grounding and intrinsic meaning. Their use of continuous vectors rather than discrete symbols does not help them escape the `merry-go-round' of representations or connect to the external world---LLMs seem trapped in a carousel of numbers instead of symbols. Like their classical counterparts, LLMs manipulate vectors that appear to lack grounding in the world and thus have no intrinsic meaning. Consequently, they encounter an analogous predicament, which we call the \textit{Vector Grounding Problem}:

\begin{displaytext}
\textbf{Vector Grounding Problem.} How can vector embeddings, manipulated solely on the basis of their numerical values according to algebraic operations, acquire meaning and refer to entities and properties in the real world---rather than merely relating to other meaningless vectors---without depending on the semantic interpretations provided by external meaning-bearing systems?
\end{displaytext}

Both versions of the Grounding Problem, as stated here, focus on the semantic content of internal states such as symbols and vector embeddings. If either problem is solved, so that their internal states acquire intrinsic meaning, we would have compelling grounds for attributing genuine semantic content to the system's outputs: the meaning of outputs would derive from the system's internal representations, rather than from post-hoc interpretations by human users.\footnote{This would create the possibility of genuine misinterpretation: human users could misconstrue the system's output by failing to grasp what the system actually represented, analogous to how humans routinely misinterpret each other's statements.}

The Grounding Problems are particularly relevant to systems that generate novel outputs. This sets them apart from systems that merely reproduce outputs generated by humans, such as paper copiers. When a human-generated text is copied, there is no question about its intrinsic meaning: it was produced by a system with meaningful internal representations to convey a specific meaning to the reader. In contrast, generative systems like LLMs produce outputs that are not, for the most part, mere copies of human-generated text; they generate novel sentences whose meanings cannot be equated to what any specific human has ever thought or said. Unlike with paper copiers, the question of whether LLM outputs are meaningful cannot be straightforwardly explained by appealing to a human author.\footnote{We will return to the question of whether the representations of LLMs only acquire extra-linguistic content through \textit{artefactual} functions that depend on the intentions of their human designers in section \ref{sec:argument-post-training}.}

\citet{Bender2020} propose a thought experiment, the `Octopus Test', to pump intuitions about the inescapability of the Vector Grounding Problem for LLMs:

\begin{displaytext}
\textbf{Octopus~Test.} Two English-speaking castaways find themselves stranded on neighbouring islands, separated by treacherous waters. Fortuitously, they discover telegraphs left by previous inhabitants, connected via an underwater cable, enabling them to communicate through telegraphic messages. Unbeknownst to them, a superintelligent octopus inhabiting these waters taps into the cable, intercepting their messages. Though the octopus has no knowledge of English, its superintelligence enables it to detect statistical patterns in the telegraphic exchanges and construct an accurate model of the statistical relationships between various signals. The octopus then severs the cable and positions itself at both ends, intercepting messages from each castaway and generating responses based solely on the statistical patterns it has learned. Whether or not the castaways detect this substitution, the octopus's messages intuitively seem devoid of intrinsic meaning. After all, the octopus merely reproduces statistical patterns gleaned from eavesdropping on human exchanges, without understanding what the signals denote. Moreover, the octopus likely fails to grasp that these signals carry meaning or serve any communicative function.\footnote{\citet{Bender2020} emphasize that the octopus lacks communicative intent when generating telegraphic signals. Since our focus is on the narrower question of vector grounding rather than the broader capacities underlying language comprehension and communication, we set aside these considerations.}
\end{displaytext}

The octopus in this thought experiment operates analogously to a powerful LLM.  Regardless of the format of its internal states---whether vectors or otherwise---the octopus accesses only linguistic signals and their statistical relationships, lacking any connection to the real-world entities and properties these signals purportedly represent. Consequently, even if the octopus produces outputs that appear perfectly sensible to human observers, these outputs lack intrinsic meaning. The Octopus Test suggests that LLMs generate outputs as devoid of intrinsic meaning as those produced by the superintelligent octopus.\footnote{One might argue that LLMs' outputs are even more devoid of intrinsic meaning than the octopus's, given that the latter is a biological organism endowed with cognitive states, motivations, superintelligence, and other features that LLMs lack. Similar considerations apply when comparing LLMs to parrots \citep{Bender2021}.  In the case of these animals, even if their outputs carry some meaning, they typically do not align with the meaning of the language they produce.} Humans (castaways or contemporary users) project meaning onto these outputs because they superficially resemble meaningful words and sentences. LLMs have no access to coconuts, trees, or any extra-linguistic reality. The vectors they process are merely arrays of numbers, grounded only in statistical patterns within language, not in the world these patterns purportedly describe. In other words, the Octopus Test invites the conclusion that all meaning in the outputs produced by LLMs is extrinsic: it lies in the eye of the beholder. This would be because LLMs are susceptible to the Vector Grounding Problem.

Throughout the remainder of this paper, we challenge this claim and the intuitions primed by the Octopus Test. We argue that, contrary to initial appearances, LLMs possess the necessary features to generate meaningful outputs based on internal representations with extra-linguistic content. Before developing this argument, we must first clarify our target notion of grounding.

\section{Five notions of Grounding} \label{fivenotions}

\begin{figure}[t]
    \centering
        \centering
        \includegraphics[width=0.6\linewidth]{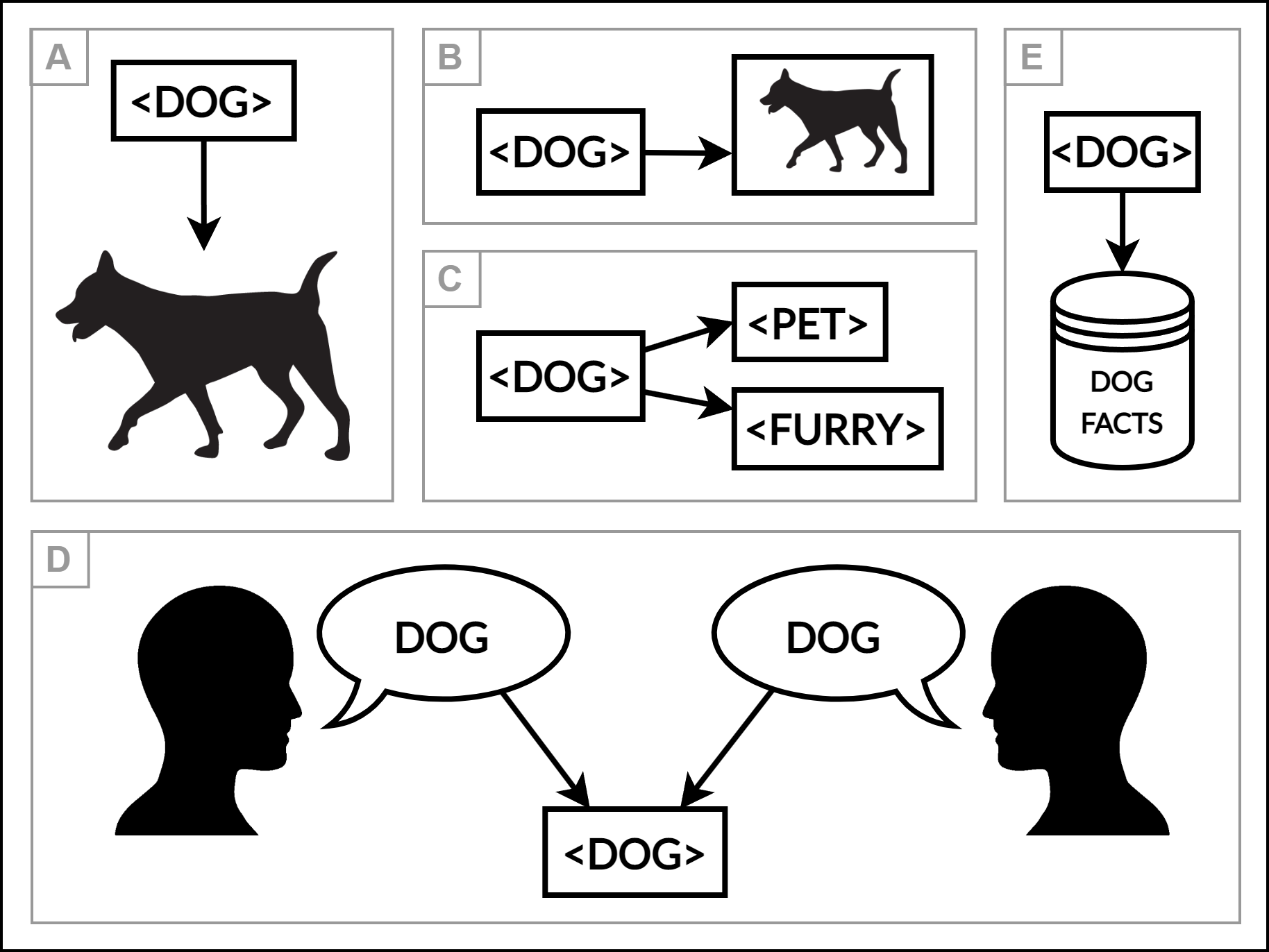}
        \caption{Five notions of grounding. A. \emph{Referential grounding}: a lexical representation (\textsc{<DOG>}) is connected to its worldly referent (a dog). B. \emph{Sensorimotor grounding}: a lexical representation (\textsc{<DOG>}) is connected to a sensory representation (image of a dog). C. \emph{Relational grounding}: a lexical representation (\textsc{<DOG>}) is connected to other lexical representations (\textsc{<PET>}, \textsc{<FURRY>}). D. \emph{Communicative grounding}: Two speakers calibrate their interpretation of an exchange to make use of the same lexical concept (\textsc{<DOG>}). E. \emph{Epistemic grounding}: A lexical representation (\textsc{<DOG>}) is connected to information stored in a knowledge base (about dog facts).}
        \label{fig:grounding}
\end{figure}

Various notions of grounding have been invoked by cognitive scientists, philosophers, and AI researchers as central for endowing internal representations and outputs with meaning, occasionally leading to confusion. In this section, we endeavour to bring clarity to this debate by identifying five primary notions of grounding at play in discussions about intrinsic meaning in cognitive systems, both biological and artificial (\autoref{fig:grounding}). Furthermore, we contend that only one of these five notions of grounding is crucial for addressing the Vector Grounding Problem. If LLMs possess features that enable them to satisfy the criteria for this particular type of grounding, they become candidates for having representations and outputs with intrinsic meaning. With this in mind, let us examine each notion of grounding in turn.

\subsection{Referential Grounding} \label{referential}

The first notion of grounding, which we call \emph{referential grounding}, closely aligns with the notion of reference. 
 It is the type of grounding that connects representations to things in the world. Reference can be loosely described as a relation that enables representations to `hook onto' worldly entities or properties.\footnote{Much philosophical work has been dedicated to examining the nature and varieties of reference. For our limited purposes, however, the intuitive notion captured by the `hooking' metaphor will suffice.} For example, the word `Paris' refers to the city of Paris, and arguably, so does a map of Paris. Both representations allow us to receive and convey information about the actual city of Paris. Similarly, many of our thoughts seem to hook onto specific entities and properties in the world, allowing us to entertain thoughts that refer to Paris, such as beliefs about it being the capital of France or desires to visit the city.

Referential grounding, as we use the term, is whatever makes it so that
representations, whether public (e.g., words and maps) or internal (e.g., beliefs and subpersonal states), hook onto the world. It is plausible that different kinds of representation require different means for hooking onto the world in the relevant sense, that is, to be referentially grounded in specific worldly entities and properties \citep{CoelhoMollo2015}.

\subsection{Sensorimotor Grounding} \label{sensorimotor}

Another type of grounding that is often invoked, particularly in
cognitive psychology, is what we might call \emph{sensorimotor grounding}. More precisely, we can divide sensorimotor grounding into two related but different views that are often conflated. \footnote{We thank an anonymous reviewer for pressing this point.}

According to what we may call the `sensorimotor representation' view of grounding, for an internal representation to be sensorimotorically grounded is for it to be fundamentally linked to sensorimotor representations available within a cognitive system.  For instance, the lexical representation `football' is sensorimotorically grounded if its representation in a
cognitive system can be reduced to, or is intrinsically connected to, sensorimotor representations of kicking movements, of the visual perception of a football pitch, of the auditory perception of the sound
of a ball being kicked, etc. This kind of grounding plays a prominent
role in theories in cognitive psychology that embrace a moderate form of
embodied cognition, such as those positing the existence of
``perceptual symbol systems'' in human cognition \citep{Barsalou1999, Pulvermueller1999}. This kind of sensorimotor grounding is the approach proposed by \citet{Harnad1990} to tackle the Symbol Grounding Problem. 

Another possible way to understand sensorimotor grounding is in terms of what we may call the `sensorimotor contact' view of grounding. Its guiding idea is that an organism's sensory and motor systems interact with the world in a distinctive, unmediated way. Specifically, sensory and motor transducers---such as nerve endings in sensory surfaces---make direct contact with the world, thus providing a basic form of grounding that is inherited, through appropriate causal chains, by sensorimotor systems and organisms as a whole.



\subsection{Relational Grounding} \label{relational}

A third notion of grounding, that we may call \emph{relational grounding}, is occasionally used to refer specifically to
intra-linguistic relationships, in accordance with the idea that a
word's meaning in a language is partly determined by its relations to other words.\footnote{The label is
  somewhat under-descriptive, since all forms of grounding are
  relational, though in different ways. What we wish to underline,
  however, is that on this sense of grounding the meaning of
  representations depends on the relations they stand in to other
  representations in the same representational system.} This  ``language-to-language grounding'' \citep{parteeMontagueGrammarMental1981} captures the way in which a
language learner can acquire knowledge about the meanings of words by
learning semantic connections between words in the language. 

Relational grounding is closely related to inferential role semantics \citep{blockAdvertisementSemanticsPsychology1986,brandomMakingItExplicit1994}, which identifies the meaning of a
linguistic expression with its inferential relations to other
expressions. There is also a clear connection between relational grounding and distributional semantics. Distributional semantics seeks to analyse word meanings in
terms of word-to-word relationships. However, instead of drawing on
linguistic analyses of word meaning, it relies on analysing co-occurrence
statistics in text corpora. Consequently, distributional semantic models
capture relationships between words grounded in patterns of use, rather
than linguistic theory. The high
dimensionality of vector spaces in LLMs makes it possible to represent complex,
multifaceted similarities between words---some of which plausibly
reflect conceptual structure and/or inferential roles \citep{Piantadosi2022}.

\subsection{Communicative Grounding} \label{communicative}

The notion of grounding is also used in relation to another aspect of
language, having to do with communication \citep{clarkGroundingCommunication1991, traumComputationalTheoryGrounding1994}. In this context, the
`grounding problem' refers to the problem of establishing \emph{common
ground} between speakers in conversation. From this perspective, linguistic communication can be seen as a form of
coordinated action that involves collaborating to reach a common
understanding of what is said. The process of communicative grounding
may involve explicit clarification strategies, such as prompting an
interlocutor to repeat or reformulate a garbled or ambiguous
statement. It also involves tracking communicative intentions during
the exchange, to infer what is communicated from what is said. Such
inferences can be made partly on the basis of pragmatic presuppositions,
namely, what speakers take for granted or assume to be true when they use
certain sentences in the context of the conversation \citep{stalnakerCommonGround2002}.

The communicative notion of grounding has also been discussed by natural language processing (NLP)
researchers \citep{brennan1998grounding, dimaroComputationalGroundingOverview2021, chanduGroundingGroundingNLP2021}, and is closely tied to the challenge to LLM meaning advanced by \citet{Bender2020}, who 
 define meaning as a relation
between linguistic expressions and communicative intent,
characterising this relation as a form of grounding.

While communicative and referential grounding are distinct notions, they
are related in important ways. Referential grounding anchors language to
the world, but the same linguistic expressions may be referentially
grounded in different ways for different speakers, due to differences in
perception, knowledge, and conceptualisation. This means that
calibrating reference in conversation requires communicative
grounding---language users must work together to negotiate a common way
of connecting language to the world. Without such collaboration, there
is no guarantee that speakers really mean the same thing, even if they
are using the same words.

\subsection{Epistemic Grounding} \label{epistemic}

Finally, the notion of grounding is occasionally used by NLP researchers
to refer more minimally to the relationship between linguistic
expressions and information stored in knowledge bases \citep{poonGroundedUnsupervisedSemantic2013,
tsaiConceptGroundingMultiple2016, chanduGroundingGroundingNLP2021,thoppilanLaMDALanguageModels2022}.
For example, an expert system may contain a knowledge base of facts and
rules belonging to a particular domain. When the system receives user
input, it can query this knowledge base to determine an appropriate
response. A classic example is MYCIN, an expert system for diagnosing
infectious blood diseases \citep{shortliffeMycinKnowledgeBasedComputer1977}. When given patient symptoms, MYCIN would
search its knowledge base of medical rules and facts to suggest possible
diagnoses and recommend treatments. In this system, language is grounded
in the structured knowledge contained in the expert system's knowledge
base. We call this notion \emph{epistemic grounding}.

There are two ways in which current LLMs may involve epistemic grounding. Firstly, they can be hooked up to external databases used to retrieve information \citep{borgeaudImprovingLanguageModels2022,thoppilanLaMDALanguageModels2022}. Secondly, it has been suggested that even LLMs that are not augmented with retrieval capabilities might be functionally similar to knowledge bases themselves \citep{petroniLanguageModelsKnowledge2019}. Indeed, they may encode a significant amount of world knowledge in their weights after pre-training, as illustrated by the remarkable performance of state-of-the-art LLMs on a wide range of factual question-answering benchmarks \citep{openaiGPT4TechnicalReport2023,bubeckSparksArtificialGeneral2023}.

\subsection{What Grounding for LLMs?} \label{groundingforllm}

We have identified five distinct notions that play
prominent roles in discussions about grounding in biological and
artificial systems. Which of these is necessary for LLMs to possess representations and produce outputs with intrinsic meaning? (Recall that intrinsic meaning is meaning that is independent from external interpretation.) To identify the most suitable notion, we will proceed by elimination, examining each candidate's adequacy---beginning with the two kinds of sensorimotor grounding.

The `sensorimotor contact' version of sensorimotor grounding claims that causal contact between sensorimotor transducers and the world grounds downstream representations. While promising, this view does not explain \textit{how} the brute impinging of light rays, sound waves and the like on sensory surfaces can ground representations in the specific worldly things they are about. Indeed, such representations are typically not about light rays or sound waves, but rather about horses, democracies, elms, and so forth.\footnote{\textit{Mutatis mutandis} for electric signals that activate muscles vis-à-vis motor plans, actions, and behaviours.} By itself, this sort of causal contact with the world does not explain how internal representations come to be about the latter, rather than (just) the former. As we will see (\autoref{theoriesrepresentation}), an account of how sensorimotor transducers can contribute to grounding requires additional components to be in place, making the sensorimotor contact version of sensorimotor grounding collapse into the notion of referential grounding: the causal contact with the world that sensorimotor transducers allow is just one of the components that can underlie referential grounding. As we will argue in \autoref{groundinglargeAI}, moreover, such unmediated causal contact with the world is not even necessary for referential grounding.

The other version of sensorimotor grounding, the `sensorimotor representation' view, connects conceptual or linguistic representations to sensorimotor representations within the same cognitive system. While this connection can illuminate the relevant representations' role within a cognitive architecture, it fails to address the merry-go-round problem. By grounding one type of representation in another, it merely postpones the crucial question: how do sensorimotor representations themselves establish contact with the world? Since sensorimotor representational grounding does not provide a connection with the world but only with other representations, it does not address the Symbol Grounding Problem. Similarly, it offers no resolution to the Vector Grounding Problem, as grounding vectors in other vectors leaves unanswered how any vector ultimately connects to worldly entities and properties.

This line of reasoning extends to relational, communicative, and epistemic grounding. Relational grounding anchors linguistic representations to their relationships to other linguistic representations. Although this kind of grounding clearly does not escape the representational carousel, it is plausible that it can suffice for endowing at least some representations with intrinsic meaning. Consider, for instance, logical connectives, for which an inferential-role semantics is compelling \citep{blockAdvertisementSemanticsPsychology1986, Fodor1990}. The meaning of a term like `and' is arguably exhausted by the role it plays in licensed inferences; it does not purport to refer to a worldly entity, but to an abstract relation between propositions. Insofar as an LLM can induce the correct inferential role for logical terms from its training data, its internal states corresponding to these terms could be said to be meaningfully grounded in that role.\footnote{Therefore, if an LLM can learn the structure of relations that constitute the meaning of logical connectives and linguistic conjunctions from its training data, then its representations of these terms might be as intrinsically meaningful as they can be.} However, logical connectives are a rather special case; at best, inferential-role semantics accounts for only a narrow subset of the representations and outputs that biological and artificial systems possess and produce. 

A similar case can be made for the representation of purely syntactic properties. The function of an internal state that represents, say, a subject-verb agreement rule or a filler-gap dependency is exhausted by its contribution to generating grammatically well-formed token sequences. Its correctness conditions are defined entirely within the formal system of the language itself. There is strong empirical evidence that LLMs, through the objective of next-token prediction alone, do in fact develop internal mechanisms that implement such grammatical rules.\footnote{See \citet{hardingOperationalisingRepresentationNatural2023c} for a discussion of criteria for ascribing representation of syntactic features to LLMs, and \citet{Milliereforthcoming} for empirical evidence that LLMs do learn such representations.}
However, the Vector Grounding Problem is not about whether a system can acquire representations of intra-linguistic structures---be they logical or syntactic---but whether it can acquire representations with extra-linguistic content. Conceding that LLMs can achieve relational grounding for formal aspects of language does nothing to solve the problem of how their representations could come to be about worldly entities and properties. Thus, while relational grounding may well be sufficient for a narrow and special subset of representations whose content is exhausted by their formal role, it remains insufficient for the vast lexicon of terms that purport to refer to the world. For these terms, relational grounding merely spins the wheels on the representational merry-go-round; it cannot, by itself, provide the traction needed to connect with extra-linguistic reality.

In sum, the form of grounding that underlies the others and directly addresses the Grounding Problems is what we have called referential grounding. Unlike the alternatives, referential grounding anchors representations in the world, rather than in other representations. It alone breaks free from the representational merry-go-round by connecting representations to whatever worldly entity or property they are about. Therefore, to determine whether generative pre-trained models like LLMs can manipulate and produce representations with intrinsic meaning, we must assess their capacity for referential grounding. This assessment requires examining what conditions must be met for a representation to achieve referential grounding.

\section{Theories of Representational Content: Causes and History} \label{theoriesrepresentation}

Philosophers of cognitive science have dedicated considerable effort over the past several decades to developing theories that explain how mental and cognitive states, such as beliefs, desires, and neural activations, can be \textit{about}, or \textit{mean}, entities and properties in the world. What mental and cognitive processes are about, or mean, is
their `representational content', or `content' for short. The philosophical project of developing theories of representational content has the aim of illuminating how meaning can emerge in a world that otherwise seems devoid of it. After all, natural objects such as rocks and lakes generally do not possess internal states endowed with meaning and, by themselves at least, do not mean anything.

Some internal representations are basic, in the sense that their meaning, or content, does not derive from other states already endowed with meaning. They can be seen as the building blocks of more complex representations, which draw some of their meaning from those basic representations. For instance, `edge detectors' in primary visual cortex (V1) are good candidates for being basic internal representations, whereas `shape detectors' in downstream visual areas are not, since their contents derive partly from the contents of the basic edge detectors. Similarly, words and sentences are not basic in this sense, since their meaning derives from other, more basic representational states, such as beliefs, desires and intentions---themselves complex representations.

The distinction between basic and complex representations cuts across the distinction between intrinsic and extrinsic meaning. Both the representations in primary visual cortex and the words and sentences humans utter are  cases of representation with intrinsic meaning, though the former are basic while the latter are complex. By contrast, words randomly formed by the movement of ants in the sand or outputs from a random sentence generator only have extrinsic meaning: they are merely shapes that require human interpretation to convey any meaning at all.

We cannot here do justice to the rich and nuanced interdisciplinary research on the notion of cognitive representation.\footnote{See \citet{Neander2017, Millikan2017, Shea2018} for recent book-length defences of the main contenders today.} What matters for our limited purposes is that, despite some disagreements on the finer details, philosophers of cognitive science have largely reached a consensus on the key factors that enable internal states in a system to become representations, both basic and complex \citep{Shea2018, CoelhoMollo2022}. 

The first, most fundamental point of agreement is that an account of basic representational content must identify one or more natural relations between internal states and external entities and properties responsible for making the former be about the latter. These natural relations must not be ones that already involve states with meaning. Since the goal is to account for the emergence of meaning in the world, theories of basic content must exclusively rely on meaning-\emph{less} states and processes to avoid vicious circularity. Much of the work in this area has been dedicated to identifying those relations that can yield fruitful content ascriptions and contribute to explaining the behaviour of cognitive systems.

Two natural relations are widely considered to be central to endowing basic internal states with meaning: \textit{causal-informational} relations, and \textit{historical} relations. In general terms, causal-informational relations obtain when one state of the world conveys information about another due to a causally-generated correlation between the two states. A classic example is the relationship between smoke and fire: smoke is correlated with fire because it is caused by it, so the presence of smoke carries information about the presence of fire. Neurons in brain area V1 convey information about edges in the perceived visual scene, as their activation correlates (through a relatively complex causal chain that includes light rays impinging on the retina) with the presence of edges in the world in front of the perceiving subject.

Causal-informational relations capture the main rationale for why systems possess basic internal representations: they allow systems to be sensitive to specific features of their environments. However, the mere existence of a causal-informational relation is insufficient to provide an account of basic representational content. An additional requirement often mentioned is that the causal-informational relation be exploited, that is, used by the system due to its information-carrying nature \citep{Shea2018}. In other words, a system must use the presence of smoke to obtain information about the presence of fire, form a belief about there being a fire, guide fire-related behaviour, and so on, for smoke to be a viable candidate for representing fire for that system. 

To be a representation necessarily involves playing a certain role: that of conveying information about the world and being used by the system in virtue of that. This further requirement complements the explanation for why internal representations are useful to systems: they are used by the system as sources of information about the world. It also helps to avoid the risk of trivialising the notion. Causal-informational relations are abundant in nature: every effect carries information about its cause. If every such relation should suffice for being a representation, representations would be so trivial that their special explanatory power for cognition would be lost.

Exploited causal-informational relations alone, however, are still not enough to base an adequate theory of representation, for another crucial feature of internal representations is missing: their capacity for truth or falsehood, correctness or incorrectness.\footnote{Correctness conditions are typical of so-called descriptive representations. Directive representations, e.g., desires and action plans, are in the business of representing how aspects of the world should be. Such representations have satisfaction conditions, i.e., the conditions that would satisfy, say, a specific desire. For simplicity's sake, we will be concerned only with descriptive representations.} To be a representation inherently involves the possibility of \emph{misrepresenting}. V1 neurons may become active when no edge is present, as in the famous Kanizsa's triangle illusion, thus conveying information to the system that, if not corrected by other subsystems, can lead to inappropriate behaviour (such as trying to touch the non-existent edge). Representations involve a kind of descriptive normativity \citep{Neander2017}. That is to say, representations can represent successfully or unsuccessfully, better or worse. This descriptive normativity extends to external representations, such as sentences, and meaning more generally. They are kinds of things that can be true or false, accurate or inaccurate, appropriate or inappropriate.
The rationale for systems to possess representations is that they be accurate enough, enough of the time. Otherwise, representations would lead systems astray and hinder appropriate behaviour, rather than the other way around. Generally, (descriptive) representations have the \emph{function} to represent correctly and accurately (to a degree of required precision).\footnote{In some special cases, systematic misrepresentation can be a sensible price to pay. If the stakes are high enough, it makes sense for representations to err on the side of caution: if you are trying to detect predators, it makes sense to produce many false-positive (misrepresenting  predators as present when there are none), rather than even one, potentially fatal false-negative.}   

As the notion of representation is central for cognitive (and computer) science, the notion of function is central to biology. This has led philosophers of biology to develop scientifically-grounded, objective theories of function \citep{Wright1973, Millik1989, Neander1991, Godfrey-Smith1994, Garson2019}. If functions exist out there in the world, they must be a matter of natural features and relations, rather than human interpretation. The dominant view of functions is the selected-effects theory \citep{Millik1989, Neander1991, Garson2019}. On this view, functions are ensembles of the historical causes that explain the persistence of a certain trait or feature in a system. For instance, hearts today have the function to pump blood because past hearts, by pumping blood, were selected due to their  contribution to the survival and reproduction of the systems having them. Pumping blood is what explains, through a historical selection process, what hearts were selected for doing---and it is thereby their function. The relevant selection processes can be at the level of the species, typically through natural selection; or at the level of the individual, through selection-like learning processes during development.

The most influential views on representation incorporate the selected-effects theory to account for internal states having the function to carry information about the world and to guide appropriate behaviour \citep{Neander2017, Millikan2017, Shea2018, CoelhoMollo2022}. The basic idea is that some states and processes in cognitive systems were selected for in virtue of their carrying information about, and guiding successful behaviour toward, states of the world---thus acquiring functions to represent such states. For example, neurons in V1 have the function of detecting edges in the visual scene because they were selected by natural selection in virtue of the causal-informational relations they maintain with actual edges in the world. When these neurons successfully detect edges, they fulfil their function, representing correctly; when they fail to fire in the presence of an edge or fire in the absence of one, they malfunction, misrepresenting. In brief, internal representations are states that carry information about the world, have the selected function to do so, and are appropriately used by the system in virtue of the information they carry.

Existing theories of basic representation vary in the emphasis they place on causal-informational relations and selection processes, as well as in the specific types of such relations they prioritise. Additionally, they may diverge on the inclusion of further relations, such as structural resemblance relations \citep{Shea2018}. Nevertheless, all of these theories, in one way or another, assert that a combination of these relations is central to the emergence of meaning in the world. To possess intrinsic meaning, at a minimum, is to stand in a specific set of causal-informational and historical relations (via past instances of selection) with the thing in the world being represented. Theories of content, therefore, provide the essential ingredients that determine referential grounding, enabling internal states to hook onto things in the world.

\section{Grounding Generative Pre-trained Models} \label{groundinglargeAI}

Philosophical theories of representation identify sets of relations that imbue representations with content, giving them intrinsic meaning. These primarily include causal-informational relations and histories of selection. Our aim is to show that generative pre-trained models, such as LLMs, can have internal states and generate outputs with intrinsic meaning, i.e., that are referentially grounded in the world.\footnote{For recent discussions that apply theories of representational content to LLMs, see \citet{hardingOperationalisingRepresentationNatural2023c,grzankowskiRealSparksArtificial2024,goldsteinDoesChatGPTHave2024, chalmersPropositionalInterpretabilityArtificial2025, williams2025structuralcorrespondencesgroundreal}.} To begin, we will examine whether appropriate causal-informational relations exist between LLM representations/outputs and entities and properties in the world.

\subsection{Causal-informational relations} \label{causalrelations}

Although LLMs are trained exclusively on linguistic data, thus precluding any direct causal-informational relationship with extra-linguistic entities, the linguistic data itself is generated by humans who stand in causal-informational relations to worldly entities and properties. Those data are shaped by humans' interactions with the world. Among its various functions, language enables humans to express facts about the world, as well as their experiences, emotions, and ideas concerning it. The extensive corpus of text upon which LLMs are trained bears the `imprint' that the world has left on language and linguistic expression.

A strict separation of form and meaning has been a central assumption in much of linguistics, from the principle of the ``arbitrariness of the linguistic sign'' \citep{Saussure1916} to contemporary formal theories of syntax and semantics \citep{chomskySyntacticStructures1957,montagueUniversalGrammar1970}. However, this notion has faced growing criticism since the 1980s, especially in light of empirical
investigation of the iconicity of language \citep{pernissBridgeIconicityWorld2014}. In this context, iconicity refers to a non-arbitrary relationship between the form of a linguistic sign and its referent, based on some similarity or analogy \citep{haimanIconicitySyntax1985}. The nature of this similarity relation can vary: imagistic iconicity involves a perceptual resemblance between a word and its referent (e.g., in onomatopoeias), while diagrammatic iconicity involves a structural resemblance.

Diagrammatic iconicity is pervasive in linguistic structures. For
example, the ordering of events in narrative sequences tends to reflect
closeness and succession in time (e.g., ``Veni, vidi, vici''); what is
nearest to the speaker in a literal or metaphorical sense tends to be
mentioned first (e.g., ``here and there''); elements whose referents are
closely related in some way or other tend to be adjacent in sentences
(principle of adjacency); the use of subordination reflects causality or
conditionality between states of affairs (e.g., ``If you study hard, you
will pass the exam''), and so on \citep{vanlangendonckIconicity2010}. The pervasiveness of diagrammatic iconicity in linguistic structures suggests that the organization and structure of linguistic signs is not arbitrary but rather reflects some ways in which the world itself is organised and structured.

While diagrammatic iconicity is not typically characterised in
distributional terms, patterns of co-occurrence in language reflect
stable statistical regularities in the way in which linguistic elements
are used together. Certain words tend to occur together more frequently
than others, and certain grammatical constructions tend to co-occur with
particular semantic categories. These patterns of co-occurrence reflect the underlying structure of the world that
speakers are trying to describe. Thus, \emph{contra} \citet{Bender2020}, form and meaning are far from being so easily separable,
especially at the level of general patterns of language use.

The patterns of language use derived from the training corpus reveal important aspects of the world. They embed information about patterns of interaction between humans and the world, human ways of categorising their environments, and the selection of features that are more or less salient to them. Indeed, a plausible hypothesis for LLMs' success in a wide range of tasks is that they have induced general semantic patterns from their training data. These patterns are not merely formal or syntactic; rather, they tap into the intricate relationships between linguistic form and the world it describes.\footnote{See \citet{merrillCanYouLearn2024} for concrete evidence that LLMs can induce entailment relationships between sentences from co-occurrence statistics.}

In short, LLMs are trained on data that is shaped by humans and their interactions with the world. LLMs stand in no direct causal connection to worldly entities, but they do stand in indirect ones, mediated by the humans who generated the training data. Importantly, while the complete lack of direct causal relations to the world sets LLMs apart from humans, reliance on indirect causal relations does not. Many human representations are also acquired by indirect, mediated ways. Indeed, testimony and social learning underlie a considerable amount of the internal representations humans come to acquire, and are considered to be key processes that help explain cultural knowledge and cultural evolution---in their turn central factors for the species' success \citep{Tomasello2009, Sterelny2014, Heyes2018}.

These include representations of entities one has not had any direct causal contact with. Most people have never been in direct causal contact with enriched uranium, but that does not entail that their thoughts and statements about uranium lack referential grounding---they are indeed about uranium. Representations of culturally constructed notions, such as `democracy' or `university', are also acquired this way---they rely on complex sets of social practices and conventions that are transmitted not by direct causal contact with documents or buildings, but rather by social learning. 

In a way akin to many human representations, the internal representations in LLMs are indirectly connected to the world through testimony and `social' learning. Since their training data contains innumerable instances of human communication, LLMs should be seen as an additional link in the chain of social information transmission that endows representations with causal-informational relations to worldly entities, and thus partly underpins intrinsic meaning.

In a sense, thereby, the causal-informational relationships between LLMs and the world are dependent on humans. This dependence is however distinct from the one that drives the Vector Grounding Problem. The Problem is not \emph{how} the representations of connectionist models come to acquire intrinsic meaning, but \emph{whether} these representations have intrinsic meaning at all---instead of being merely ascribed meaning by external interpreters. LLMs find themselves in a position similar to most humans: their representations and outputs about, say, enriched uranium intrinsically refer to enriched uranium---irrespective of any interpreter's gloss---even though the process enabling such reference depends on a chain of transmission that involves other systems with intrinsic meaning. In our terminology, such internal states are non-basic representations with intrinsic meaning.

In sum, LLMs possess the necessary elements to fulfil the causal-informational requirement for referential grounding. They stand in complex, mediated causal-informational relations to the world. These connections are mediated since they pass through human linguistic production, where language primarily functions as a tool to express aspects of the world.\footnote{This analysis would extend, \emph{mutatis mutandis}, to multimodal inputs, which are also mediated by human interactions with the world (e.g., photographs are captured by humans with a camera). We discuss whether multimodal models fulfil the historical requirement for representational content in \autoref{multimodal-embodied} below.} They are complex due to the intricate causal-informational relationships between the patterns of language use, the structure of the world they have (culturally) evolved to capture, and the causal chains of communication and testimony that produce the learning data---of which the LLM itself can be considered a link. In light of these considerations, we contend that generative pre-trained models, such as LLMs, can meet the causal-informational criteria necessary for possessing referentially grounded representations and outputs.

\subsection{Selection history} \label{historicalrelations}

Finding causal-informational relations between LLMs and the world is the most straightforward part of our investigation, given the abundance of such relations. The far greater challenge lies in showing that LLMs have internal states that not only carry information about worldly entities and properties, but also have the \textit{function} of carrying such information---enabling LLMs to generate appropriate outputs based on this representational content.

Theories of representation typically invoke selection processes, broadly construed to include learning, as the historical mechanisms that explain how internal states acquire representational functions.
While generative pre-trained models clearly undergo a form of learning during training, whether this constitutes the appropriate kind of selection process is less clear. The crucial question is whether LLMs' internal states possess the \emph{right} kind of selection history---having been selected for carrying information about something and guiding behaviour accordingly---directed toward the \emph{right} targets, namely the entities and properties in the world that could ground their representations \citep{Butlin2021}. Only when these conditions are satisfied can LLMs' internal states acquire genuine representational functions and the descriptive normativity that accompany them.

It is important to distinguish between the \textit{proximate} and \textit{ultimate} functions that LLMs perform during training and inference. The proximate function of LLMs is plainly next-token prediction: given a text sequence, predict the subsequent token based on contextual information. This proximate function is intra-linguistic, and so is its success criterion---namely, the LLM's accuracy at predicting the correct tokens. While this function may support intra-linguistic \emph{relational grounding}, it appears insufficient for establishing the normativity required for genuine reference to the external world. Ultimate functions, by contrast, encompass the broader, more abstract goals that models achieve through next-token prediction. These functions can, in principle, involve world-directed tasks such as generating text that accurately describes real-world states of affairs or solving problems requiring real-world knowledge. We will examine several potential sources of world-involving, representational ultimate functions in LLM training and deployment.\footnote{The relevant world need not be our shared, physical world, but can encompass other extra-linguistic domains, such as the `game world' of chess rules, pieces, and moves. When discussing this more liberal notion of `world-involving', we note it explicitly in the text.} 

Our general argument goes as follows. The training process of LLMs is a form of selection through learning that picks out and stabilises those internal states that lead to successful behaviour. In many cases, this process selects for internal states whose ultimate function extends beyond mere linguistic prediction to include extra-linguistic, world-involving correctness conditions, such as representing facts about the world. This is most evident in models that undergo fine-tuning to satisfy human preferences, since such fine-tuning explicitly involves selecting internal states that increase the probability of outputs satisfying world-involving norms, such as factual accuracy. We will also investigate the possibility that, to a limited extent and within some constrained domains, pre-training on next-token prediction alone may be sufficient for endowing internal states with world-involving, representational functions. If LLMs indeed have internal states with extra-linguistic intrinsic meaning, and if those states are causally responsible for the generation of linguistic outputs, then those outputs are thereby referentially grounded in the world. This would mean that at least some outputs of LLMs are intrinsically meaningful in the sense discussed earlier: their meaning does not depend on the interpretations projected onto them by human users.

The key premise in this argument---that the training LLMs undergo is a kind of selection process suitable for endowing their internal states with functions to represent states of the world---requires careful defence. In what follows, we explore three complementary strategies to defend the premise, each aimed at showing that a specific kind of learning process in current LLMs can in principle confer world-involving, representational functions to their internal states. The \textit{argument from post-training} focuses on fine-tuning methods based on human preferences as the most promising candidates for this role. The \textit{argument from pre-training} investigates whether the basic next-token prediction objective might, under specific conditions, implicitly select for world-involving representational functions. Finally, the \textit{argument from in-context learning} proposes that LLMs could, in principle, acquire world-involving representational functions during inference without parameter updates.

\subsubsection{The Argument from Post-Training}
\label{sec:argument-post-training}

The training process of modern LLMs, particularly those deployed in consumer-facing chatbots like ChatGPT, involves two distinct phases: \emph{pre-training} and \emph{post-training}. Pre-training involves training the base model from scratch on a next-token prediction objective across vast corpora of text. By contrast, post-training adapts this base model to behave in ways that go beyond generating merely plausible text completions, making LLMs more useful for various downstream tasks and applications. This phase typically includes two types of fine-tuning: \emph{instruction tuning} and \emph{preference tuning}.

Instruction tuning is a form of supervised fine-tuning on curated instruction-response pairs \citep{ouyangTrainingLanguageModels2022}. Given an instruction, the model is trained to generate the desired output by minimising the difference between its predictions and the target responses in the instruction dataset. This teaches the model to generate appropriate responses that follow user instructions rather than simply generating plausible completions of text sequences. Preference tuning further trains LLMs to generate outputs that better align with selected normative criteria, such as helpfulness, harmlessness, and honesty or factual accuracy \citep{askellGeneralLanguageAssistant2021}. Unlike standard supervised fine-tuning, which only demonstrates what to do, preference tuning explicitly trains models to distinguish between better and worse responses according to these criteria. This is typically implemented by training on pairs of responses where human evaluators have indicated a preference for one over the other. Reinforcement Learning from Human Feedback (RLHF) is the most well-known preference tuning method \citep{christianoDeepReinforcementLearning2017}. RLHF begins by collecting human preferences over model outputs for a diverse set of prompts. These preferences are used to train a reward model that predicts how humans would evaluate various responses. The LLM is then fine-tuned using reinforcement learning to maximise this reward function while remaining reasonably close to its original behaviour. Another increasingly popular approach is Direct Preference optimisation (DPO), which achieves similar results without explicitly modelling the reward function \citep{rafailovDirectPreferenceOptimization2024}. Instead, DPO directly optimises the model to assign higher probabilities to preferred responses and lower probabilities to rejected ones.

When an LLM is optimised according to human preferences, a selection process is established that reinforces specific internal states based on their contribution to externally rewarded outputs. Consider a toy example using DPO. We begin with a language model that has undergone both pre-training and instruction tuning. This model can follow instructions reasonably well, but has not been specifically optimised for factual accuracy. To align the model with the norm of factual accuracy, we collect a dataset of human preferences where judges rank responses based solely on which one contains more truthful information and fewer falsehoods. When we apply DPO to this model, it adjusts the model's parameters to increase the probability of generating factually accurate responses and decrease the probability of generating less accurate ones. Through this process, the model's internal states are selected because they enable the model to satisfy the norm of factual accuracy, thereby providing a standard of success that depends not on linguistic patterns alone, but on the correspondence between the generated text and states of affairs in the world. This provides the missing piece of the story---in addition to causal-informational relations---of how the internal states of LLMs can be referentially grounded, and thus overcome the Vector Grounding Problem. The internal states of the preference-tuned LLM represent states of the world because they were selected specifically for their ability to carry information about those states in service of generating truthful outputs. When the model succeeds in generating a factually accurate response, its internal states fulfil their function of representing relevant worldly states. 

Two clarifications are in order. Firstly, we should distinguish between the function of the learning mechanism and the function of the representational states selected by that mechanism. The designers' intentions bestow an artefactual function on the learning mechanism (e.g., the preference tuning pipeline has the intended function of updating parameters based on reward signals). However, the function of the model's internal states is set by the learning process itself. The situation is analogous to artificial selection in biology. A dog breeder's intention to produce calmer dogs establishes a selection regime, but it is the objective history of differential reproduction that confers the biological function of producing calmness on the selected traits. This function is a fact about the lineage's history, not the breeder's mind. Similarly, the designers of an LLM create a selection apparatus, but the autonomous operation of this apparatus on data confers new, selected-effects functions on the model's internal states. The normativity---the standard for correctness or incorrectness---is supplied by this objective history of selection, not the designers' intentions. 

Secondly, the extra-linguistic content of the model's representations is determined by the specific worldly conditions that drove their selection, which in the case of preference tuning means whatever properties the reward signal actually and reliably tracks. In that regard, the focus on factual accuracy in our example is a special case. When human evaluators consistently provide higher rewards for factually accurate statements, the selection process picks out and stabilises the model's internal states that carry relevant factual information about the world, in virtue of their doing so. The other two common norms in preference tuning---helpfulness and harmlessness---are also suitable to selecting referentially grounded internal states. Helpfulness requires generating outputs that genuinely assist users with their goals, which hinges on real-world task success conditions rather than mere linguistic coherence. Harmlessness involves avoiding outputs that could lead to real-world negative consequences, requiring sensitivity to causal relationships between linguistic outputs and potential worldly outcomes. Even instruction tuning alone could establish world-involving functions, as successfully following instructions often requires tracking aspects of the world that would make the instruction completion successful. Many instructions implicitly contain success conditions that depend on extra-linguistic factors. 

In summary, post-training establishes the selection processes necessary for endowing LLM internal states with representational functions, which together with causal-informational relations are the key ingredients for intrinsic meaning. It is worth noting that the specific norms used in post-training shape what the internal states represent. Achieving referential grounding does not automatically guarantee that the meanings of the internal representations and outputs of LLMs align with their conventional human meanings. Un-humanlike selection criteria, together with the other un-humanlike features of LLMs, may endow them with `deviant meanings'. This is an intriguing possibility that lies beyond the scope of this paper. What matters for our purposes is that LLMs' internal states and outputs can be intrinsically meaningful, even if such meanings may be alien to humans.    


\subsubsection{The Argument from Pre-Training}
\label{argument-from-pre-training}

We now turn to a more controversial claim: that pre-training alone can, under specific conditions, establish world-involving representational functions. The proximate function of an LLM during pre-training is to minimise the negative log-likelihood of the next token given a preceding sequence.\footnote{See \citet{McCoy2024} for an approach to studying LLMs, which they call the ``the teleological approach'', that takes as its starting point this proximate function. They show that LLM behaviour is strongly shaped by the distributional statistics of the training dataset. As we argue here, however, this is compatible with LLMs also, at least in some cases, fulfilling ultimate, world-involving functions. Sensitivity to distributional patterns can be a means to fulfilling ultimate goals, given all the worldly information embedded in such patterns.} At first glance, this appears to be a purely intra-linguistic objective, with correctness conditions defined entirely by the statistical distribution of tokens in the training corpus. It is not immediately obvious how a selection process governed by such a criterion could give rise to internal states whose ultimate function is to represent extra-linguistic reality.

To succeed at next-token prediction, two strategies are possible: memorising observed sequences or inducing latent factors responsible for those sequences. For large datasets, the latter strategy is far more efficient \citep{deletangLanguageModelingCompression2024}. This can create a selection pressure for internal states that track underlying causal factors, as they enable more efficient prediction across diverse linguistic contexts. Such selection pressure may be enough to endow language models with internal states that stand in mapping relations to states of affairs in the world, sometimes called `world models' \citep{yildirimTaskStructuresWorld2023}. However, philosophers of representation have long recognised that standing in mapping relations to states of the world---also known as `structural resemblance'---is insufficient for representation. Such mappings are cheap: they can be found between any two sets with the same cardinality. Representational content would be trivial and ubiquitous if structural resemblance were sufficient to secure it \citep{Shea2012-SHEMIR}. For internal states that merely map onto extra-linguistic domains to become full-fledged representations, they must have the function of representing those worldly domains and be used by the system to generate outputs precisely because they have this representational function \citep{Shea2018, williams2025structuralcorrespondencesgroundreal}.

The concept of meta-learning provides a formal framework for understanding how pre-training could, in principle, establish world-involving representational functions. Meta‑learning, or `learning to learn', is a bi‑level optimisation process with two learning loops: the outer loop trains a model over a range of related tasks, while the inner loop refers to the model's rapid adaptation when it encounters data from a specific task. Rather than seeking perfect performance on any one training task, the outer loop shapes the model’s parameters so that a small number of gradient steps (the inner loop) can quickly tailor those parameters to a new, previously unseen task from the same distribution. The pre-training of LLMs can be viewed as an implicit form of meta-learning \citep{wuMetaLearningPerspectiveTransformers2024}. The outer loop is the standard training process: minimising next-token prediction loss via gradient descent on the model's parameters over a large corpus. This corpus implicitly contains a massive distribution of distinct tasks---translating, answering questions, summarising, solving problems, playing games, etc. The outer loop thus selects for a single set of parameters that makes the model an effective learner for the myriad tasks embedded in its training data. Wu et al. suggest that the inner loop is the forward pass itself. On their analysis, the fixed parameters learned during the outer loop do not encode a static knowledge base, but rather an on-the-fly optimisation algorithm. For any given input sequence (i.e., a task context), the forward pass through the Transformer's layers functions as an iterative optimisation process. Each layer takes the token representations from the layer below and performs an update, which Wu et al. show can be mathematically interpreted as an optimisation step that refines the representations to make them progressively more suitable for the final next-token prediction.

This perspective provides a plausible, though speculative, mechanism for the emergence of world-involving representational functions during pre-training. The outer loop's general, domain-agnostic objective (predict the next token) exerts a selection pressure on the dynamics of the inner-loop optimiser (the forward pass). For specific domains represented in the training data, the most efficient strategy for the inner loop to succeed at its task-specific prediction objective may be to construct and manipulate an internal model of the latent, extra-linguistic state of affairs that generated the data. The proximate, intra-linguistic objective of the outer loop would thereby select for an inner-loop process whose success is conditional on tracking extra-linguistic states. In such cases, the resulting internal states would acquire the ultimate function of representing that extra-linguistic domain. Their function would be to carry information about the world because their having been selected by the learning process is explained by the fact that they did so, thereby enabling more accurate predictions.

The argument's viability rests on two substantive assumptions. The first is that the meta-learning perspective provides a valid mechanistic account of the Transformer's forward pass during pre-training. The second is that for certain domains, the general pre-training objective (next-token prediction) creates selection pressures for an inner-loop process whose success is conditional upon tracking extra-linguistic objectives. The argument is most plausible for what we might call \textit{formally constrained domains}, such as logic puzzles or board games. In such domains, the condition specified in the second assumption is most clearly met. The sequences of tokens in the training data (e.g., game moves) are not generated by the contingent communicative intentions of humans in an open-ended world, but are instead governed by objective, formal rules that constitute the domain itself. The statistical regularities in the data are direct consequences of this underlying formal structure. For a model to succeed at its proximate, intra-linguistic task of predicting valid next tokens, it must become sensitive to these non-accidental regularities. The most robust and efficient strategy for achieving this is to induce an internal model of the domain’s rules and states---the very structure that generates the data. This establishes a non-arbitrary, world-involving standard of correctness, allowing the inner-loop selection process to endow internal states with the ultimate function of representing that extra-linguistic domain.

Research in mechanistic interpretability provides concrete evidence that models pre-trained on next-token prediction can develop internal representations with extra-linguistic content at least in formally constrained domains. One particularly interesting case is Othello-GPT, a transformer model trained solely to predict moves in Othello game transcripts \citep{liEmergentWorldRepresentations2023}. Despite receiving no explicit information about Othello's rules or board structure, this model was found to develop internal representations that encode the current game state in a player-centric manner. \citet{nandaEmergentLinearRepresentations2023} showed through intervention studies that specific directions in the model's activation space encode board positions as `Mine', `Yours', or `Empty'. When they intervened on these representations by `flipping' a tile from `yours' to `mine' in the model's activation space, the model's predictions changed in ways consistent with the altered board state. This suggests not only that the model encodes the board state, but that this encoding plays a causal role in determining its predictions of legal next moves.\footnote{For a longer philosophical discussion of causal interventions used to understand internal states of Othello-GPT, see \citet{milliereInterventionistMethodsInterpreting2025}.}

The Othello-GPT example illustrates how selection processes during pre-training may establish representational functions with extra-linguistic correctness conditions. During training, the model developed specific directions in its activation space that linearly encode the board state from the perspective of the current player. These linear representations plausibly emerged and persisted because they enabled the model to accurately predict legal moves across diverse game contexts. Their function thus became that of carrying information about the current state of the Othello board, with accuracy conditions dependent on the actual game state and the learned rules, rather than merely on linguistic patterns. This example is particularly compelling because game transcripts are systematically constrained by the rules of Othello---to predict legal moves effectively, the model developed a linearly decodable internal model that tracks the underlying game state.\footnote{Further research suggests that Othello-GPT may learn many independent, localised decision rules (a ``bag of heuristics'') rather than a single systematic algorithm for tracking the board state and determining legal moves \citep{jylin04OthelloGPTLearnedBag2024}. For instance, researchers identified specific ``board pattern neurons'' that activate when a particular local arrangement of pieces makes a specific move legal. This is consistent with our argument; the crucial point is that the internal states guiding such heuristics acquire a world-involving representational function through their selection history. A given ``board pattern neuron'' was selected by the training process \textit{because} its activation reliably correlated with a specific, rule-governed state of affairs on the Othello board, and detecting this state of affairs improved the model's ability to predict a legal next move. The correctness condition for this neuron's firing is therefore extra-linguistic: it is supplied by the actual state of the Othello board as governed by the game's objective rules. If the neuron fires when the pattern is not present, or fails to fire when it is, it has malfunctioned with respect to the game of Othello, and this malfunction increases the probability of being penalised by the loss function.}

The case of LLMs trained on diverse corpora is undoubtedly more complex than Othello-GPT. Games represent a special class of domains with respect to the grounding problem. Unlike open-ended conversation, where success criteria may be fuzzy and context-dependent, games typically have explicit, formalised rules that define success and failure categorically. These rules create a clearly bounded domain with a fixed ontology (game pieces, positions, legal moves), and deterministic success conditions. By contrast, general LLMs encounter messy, open-ended data with few rigid underlying rules. To the extent that language models may acquire internal states with world-involving content purely through pre-training, similarly to Othello-GPT with respect to the game world, this presumably only happens in formally constrained domains, in which next-token prediction on text sequences provides the right kind of learning signal---provided that the two aforementioned assumptions hold.

In summary, the argument from pre-training suggests that even without post-training methods like preference tuning, language models may develop internal representations with ultimate extra-linguistic representational functions through the selective process of next-token prediction over structured text data, at least in some limited, formally constrained domains. Evidence from mechanistic interpretability studies supports this view at least in simple game domains, showing that generative pre-trained models can develop internal states that carry information about extra-linguistic properties and causally influence outputs.

\subsubsection{The Argument from In-Context Learning}
\label{argument-from-in-context-learning}

After pre-training, LLMs can rapidly adapt to new tasks from just a few examples at inference time, a capacity known as \textit{in-context learning} (ICL). Through ICL, pre-trained LLMs appear capable of `learning' various tasks on the fly without any update to their internal parameters. These tasks can be intra-linguistic (e.g., translating between languages) or extra-linguistic (e.g., answering questions about world capitals). An intriguing question is whether ICL can be sufficient to achieve referential grounding.

Suppose that we invent a new board game called `Chroma', played by placing coloured tiles on a grid. The game has a clear set of rules: for instance, a blue tile can only be placed adjacent to a red or yellow tile, and scoring depends on creating specific geometric patterns. These rules define the game's `world'---a self-contained domain with its own entities (tiles, grid positions), properties (colours), and causal regularities (legal moves, scoring conditions). Now, imagine we present a pre-trained LLM, which has never encountered Chroma before, with a prompt containing the complete rules of the game, followed by a transcript of the first few moves. We then prompt it to generate the next legal and strategically optimal move. For the LLM to succeed at this task, it cannot rely on merely matching statistical patterns from its pre-training data, as Chroma is a novel game. It must, in some sense, induce and apply the rules of the game without further training. To the extent that an LLM can do this through ICL, we argue that this can be understood as a selection process that establishes world-involving representational functions for the model's internal representations.

Recent research on ICL suggests that in formal domains, Transformers can learn to use the in-context examples to construct an implicit, task-specific objective function and then perform computations functionally equivalent to gradient descent to find a solution---a process called `mesa-optimisation' \citep{oswaldUncoveringMesaoptimizationAlgorithms2024}.\footnote{Mesa-optimisation at inference time is analogous to the aforementioned inner-loop optimisation during pre-training.} The initial layers of the model parse the context and construct what is effectively a training set within the model's activation space. Subsequent layers then iteratively refine the internal representations, taking steps to minimise an implicit loss function tailored to the task at hand. During ICL, this mesa-optimisation process within the forward pass constitutes a selection process that provides a transient locus of normativity for the system's internal states, even without parameter updates. It selects for internal activation patterns because they contribute to satisfying the task function defined in context. In the Chroma example, this task function is plausibly extra-linguistic with respect to the game-world. The objective that would have to be mesa-optimised is not to predict a linguistically plausible next token, but to identify a move that is valid according to the game rules. In other words, the correctness conditions for this task are determined not by the statistics of English, but by the objective state of the Chroma game world. 

For example, an activation pattern that correctly represents that ``placing a blue tile at C4 is a legal move'' will be reinforced and stabilised during the forward pass because it helps to minimise the implicit error against the objective of making a valid move. Conversely, a pattern representing an illegal move will be suppressed. These internal states thereby acquire a transient selected-effects function: their function is to carry information about the state of the Chroma game (e.g., the board configuration, legal moves, strategic opportunities) in service of winning the game. This function, and the normativity it entails, is established on-the-fly, for the duration of solving the specific task presented in context, and will cease to exist once that context is deleted or overwritten.

Whether this actually occurs in LLMs is an open question. The emergence of effective mesa-optimisers is not guaranteed and depends on the properties of the pre-training data distribution \citep{zhengMesaOptimizationAutoregressivelyTrained2024}. Nonetheless, the argument from in-context learning provides a speculative but mechanistically plausible route by which an LLM could, in principle, acquire short-lived representations with extra-linguistic content on the fly.


\section{Implications}
\label{sec:implications}

\subsection{Parameter-identical duplicates}

The historical component of referential grounding invites consideration of a classic philosophical objection. The `swampman' thought experiment envisions a perfect duplicate of a human spontaneously formed by random chance (in the original thought experiment, by a lightning strike on a swamp), and thus without any causal history. We can construct an analogous thought experiment for LLMs:

\begin{displaytext}
\textbf{Swamp LLM.} Suppose a language model $M_1$ is trained normally, undergoing both pre-training on next-token prediction and post-training with instruction tuning and preference tuning. Now imagine that an untrained language model $M_2$ is randomly initialised and, by extraordinary coincidence, happens to have exactly the same parameter values as the fully trained $M_1$, despite never having been exposed to any training data.
\end{displaytext}

If $M_1$ achieves referential grounding on our view, does $M_2$ also achieve it? Our answer aligns with the standard response to the swampman scenario: despite being parameter-identical to $M_1$, $M_2$ would lack referential grounding. First, the internal states of the swamp system, at least at the moment of creation, would lack any causal-informational relations to the world, given that, by hypothesis, the system has no causal history. Moreover, the absence of a selection and learning history means that the internal states of the swamp system lack any functions, including functions to carry information about the world \citep{Shea2018}. Consequently, there is no principled way to determine which outcomes count as successes or failures for a swamp system. $M_2$ might produce outputs that appear meaningful to human interpreters, but without the selection history that shaped $M_1$, there is no objective basis for treating some of its outputs as `correct' and others as `incorrect'. The parameter values themselves cannot ground representational content.

It is worth noting, however, that referential grounding could be established relatively quickly if $M_2$ were subsequently fine-tuned. Even a brief learning history would begin to establish the relevant world-involving functions \citep{Shea2018}. What matters for grounding is not the parameter values per se, but the historical process through which those parameters were selected to perform specific functions.\footnote{A subtle variant of this case is a Swamp LLM with frozen weights that would nonetheless engage in in-context learning, adapting its outputs based on information provided within a conversational context. One might be tempted to argue that this `inner loop' of adaptation within the forward pass constitutes a sufficient selection process to establish world-involving functions \textit{ab initio}. We contend that it does not. The capacity for ICL is itself a complex functional property that requires a selection history. In a normally trained model, this capacity is the product of the `outer loop' of pre-training, which selects for internal mechanisms that can efficiently adapt to the diverse tasks and statistical patterns encountered in the training data. The ICL mechanism in such a model thus has a selected-effects function: to enable rapid task adaptation based on contextual cues. A Swamp LLM, by contrast, possesses this mechanism by chance, so the mechanism itself lacks any function. Its apparent adaptation is a form of pseudo-learning---a causal process that mimics genuine, normatively-grounded adaptation but lacks the historical basis for correctness conditions. Without a function to represent correctly, its internal states cannot misrepresent, and therefore cannot be said to represent at all. We are grateful to Harvey Lederman for suggesting this example.}

We can also consider a variant of this thought experiment:

\begin{displaytext}
\textbf{Lucky Pre-Training.} Suppose $M_1$ is trained normally, undergoing both pre-training and post-training. $M_2$ undergoes only pre-training on next-token prediction, but through some extraordinary coincidence, ends up with exactly the same parameter values as the fully trained $M_1$.
\end{displaytext}

\noindent This scenario differs importantly from the standard swampman case, as $M_2$ does have a learning history---it just differs from that of $M_1$. The answer to whether $M_2$ achieves referential grounding depends on whether one accepts our argument from pre-training in Section \ref{argument-from-pre-training}. If one accepts that pre-training alone can, in some cases, establish world-involving representational functions, then $M_2$ could achieve referential grounding despite skipping the post-training phase. The fact that $M_2$ happens to have identical parameter values to $M_1$ would be a mere coincidence irrelevant to the grounding question. Indeed, its referential grounding would be limited to that achieved via pre-training, while the internal states that drew their grounding from preference tuning in $M_1$ would be ungrounded in $M_2$.

These thought experiments highlight the crucial role that a history of selection plays in referential grounding. The internal states of a system, no matter how complex or functionally organised, cannot have representational content without the appropriate history. At the same time, our position avoids an overly restrictive view of which learning histories can establish referential grounding, allowing that different training regimes might enable a system to overcome the Vector Grounding Problem through different causal-historical paths.

\subsection{The impact of fine-tuning}

In a recent paper, \citet{grindrodLargeLanguageModels2024} raises an objection to our argument from post-training. Grindrod acknowledges our claim that fine-tuning with human feedback introduces selection processes that could in principle establish the necessary historical relations for referential grounding. However, he objects:

\begin{quote}
The problem with this idea is that it would be odd to claim that a relatively marginal procedure that is designed to optimise the performance of a LLM on specific tasks ends up being the difference between intentionality and its lack. The reason, simply put, is that the fine-tuning procedures---whether RLHF, supervised training, or something else---are not the reason why we have such impressive LLMs today. \citep[][p. 71]{grindrodLargeLanguageModels2024}
\end{quote}

\noindent Grindrod's objection rests on the premise that it would be implausible for a ``relatively marginal procedure'' to be the critical factor determining whether LLMs possess intrinsic meaning. He points out that pre-trained models without preference tuning already demonstrate impressive capabilities, suggesting that whatever makes LLMs impressive cannot be the same factor that provides them with intrinsic meaning.

This objection misunderstands the purpose of our appeal to preference tuning. We are not claiming that preference tuning explains the impressive capabilities of LLMs or their ability to produce contextually appropriate outputs. Rather, we are specifically addressing how LLMs can overcome the Vector Grounding Problem---how their internal states can possess intrinsic meaning by being referentially grounded in the world. These are distinct questions. Intrinsic meaning does not determine the complexity or impressiveness of outputs, neither in biological nor in artificial systems. Many relatively simple biological systems possess internal states with intrinsic meaning, while many complex systems might generate impressive outputs without intrinsic meaning. The quality of outputs and their status as intrinsically meaningful are orthogonal issues. What matters for referential grounding is not how much a particular training procedure contributes to performance quality, but whether it helps to establish the right kinds of relations between internal states and worldly entities. Even if preference tuning procedures make only marginal contributions to an LLM's overall capabilities, they can fundamentally alter the normative conditions under which the model operates by introducing world-involving correctness standards.

\subsection{Multimodality} \label{multimodal-embodied}

At first blush, multimodal and embodied AI systems seem to be the most obvious candidates for overcoming the Vector Grounding Problem. Unlike pure LLMs, they receive sensory inputs (e.g., pixels) and/or interact physically with the world, so that the sort of `sensorimotor contact' we discussed in \autoref{sensorimotor} seems to obtain. In \autoref{theoriesrepresentation}, we pointed out that an account of grounding based on mere causal contact is incomplete. Indeed, it follows from our analysis in \autoref{groundinglargeAI} that multimodality and embodiment are neither necessary nor sufficient for referential grounding, though they can be part of what makes referential grounding obtain when combined with the right kind of learning process.

Let us first consider a simple multimodal model like CLIP \citep{radfordLearningTransferableVisual2021}. CLIP is developed by jointly training an image encoder and a text encoder to predict the correct pairings of images and text captions using a contrastive learning objective. During training, CLIP receives batches of image-text pairs and learns to maximise the similarity between embeddings of matched pairs while minimising the similarity between embeddings of unmatched pairs. As far as causal-informational relations go, the account we gave for LLMs should also apply to models like CLIP: since both images and their captions are shaped by human interactions with the world, CLIP should stand in complex, mediated causal-informational relations to the world. The images in CLIP's training data were produced by humans using cameras to capture real-world scenes, and the captions were written by humans to describe those scenes. This establishes an indirect causal chain connecting CLIP's internal states to worldly entities and properties. 

However, the contrastive learning objective optimises purely for matching images with their captions, without establishing world-involving functions for CLIP's internal states. There is no obvious selection pressure for internal states that track features of the world beyond what is necessary for the matching task, and no clear normative standard for success beyond the matching task itself.  Consequently, CLIP lacks the selection history necessary to imbue its states with world-involving representational functions. It therefore fails to achieve referential grounding.

By contrast with CLIP, some multimodal models are built with a pre-trained language model backbone. For example, typical VLMs that can receive images as well as text as input, such as Flamingo and LLaVA, leverage existing LLMs (with frozen weights) and learn a mapping between the visual encoder and the LLM \citep{bordesIntroductionVisionLanguageModeling2024}. This approach is computationally more efficient than training both vision and text encoders from scratch, as it only requires learning the connection between modalities. To the extent that the LLM backbone achieves referential grounding from pre-training on text alone, our framework entails that such a model should inherit this property. The addition of a superficial mapping to the embedding space of a visual encoder provides a form of sensorimotor grounding, but it does not contribute to the referential grounding inherited from the pre-trained LLM. 

\subsection{Embodiment} \label{embodiment}

Embodiment is often mentioned as the best way to ensure that AI systems are grounded in the world, via their robotic bodies. Some embodied systems like SayCan \citep{ahnCanNotSay2022} integrate a pre-trained LLM with a robotic body in a way that keeps these components functionally separate. In SayCan, an LLM breaks down verbal instructions into sequences of `atomic' behaviours that the robot can enact through low-level visuomotor control. Importantly, the LLM component is not further trained or fine-tuned based on the success or failure of the robot's actions---there is no selection of internal states on the basis of their contributions to successful behaviour. In such systems, the embodied aspect adds nothing to the referential grounding of the language model component. Without additional training that would establish new causal-historical relations between the LLM's internal states and the outcomes of physical actions, the LLM's capacity for referential grounding remains exactly what it was before integration with the robotic system. Like multimodality, physical embodiment alone does not automatically contribute to referential grounding. 

When fine-tuning is involved, the story is a bit different. Recent advances in robotic learning have led to the development of sophisticated multimodal systems that integrate vision, language, and action. Models such as RT-2 \citep{zitkovichRT2VisionLanguageActionModels2023a}, RT-X \citep{oneillOpenXEmbodimentRobotic2024}, and OpenVLA \citep{kimOpenVLAOpenSourceVisionLanguageAction2024} typically build upon language models or vision-language models that were pre-trained on text or multimodal data before being fine-tuned for robotic control. These models incorporate a pre-trained language model component and may inherit referential grounding from it. To the extent that the underlying language model achieves referential grounding through the mechanisms we have identified, the system as a whole may inherit this grounding.

However, VLA models go beyond merely inheriting referential grounding from their language model components. During embodied training, they are specifically trained to map visual inputs and natural language instructions to robotic actions that achieve specific goals in the physical world. This additional training process may endow the model's internal states with world-involving, representational functions grounded in a history of learning through physical interactions with the world.

\section{Concluding Remarks} \label{conclusion}

We have argued that the Symbol Grounding Problem, as introduced by \citet{Harnad1990}, also applies to deep neural networks such as generative pre-trained models, albeit in a slightly different form, which we have called the Vector Grounding Problem. We have sought to demonstrate that at least some current generative pre-trained models satisfy the requirements for having representations and outputs grounded in extra-linguistic referents.

While our discussion has focused on LLMs and, to a lesser extent, on their multimodal counterparts, any AI system that fulfils the causal-informational and selection history requirements delineated above can possess representations with intrinsic meaning. Meeting these requirements is far from trivial, but nonetheless within reach of current AI models. As philosophical theories of representation have shown, to understand how cognitive systems---both biological and artificial---function, we must resist assuming that intrinsic meaning necessarily entails complex cognitive abilities such as understanding, knowledge, or consciousness. While having states imbued with intrinsic meaning might be a fundamental condition for possessing these cognitive abilities, it is insufficient on its own. Activations of specific populations of neurons in the mammalian cortex may represent the presence of edges in the visual scene, but these neuronal activations do not understand or possess knowledge about edges. We suggest that a similar claim can be made regarding the internal representations and outputs of LLMs and related connectionist models.

We argued for this claim in two steps. First, we distinguished between various notions of grounding often conflated in the cognitive science and computer science literature. We suggested that one particular notion, \emph{referential grounding}, is central to both the Vector Grounding Problem and the older Symbol Grounding Problem. Referential grounding pertains to how representations hook onto the world. It is the most fundamental type of grounding, and arguably serves as the basis for all other notions of grounding (with the partial exception of relational grounding). Referential grounding requires two main factors: causal-informational relations between internal states and the world, and a selection history that imbues those states with functions to carry information about the world.

Second, we have shown that certain LLMs trained exclusively on linguistic data, particularly those that undergo preference tuning, are referentially grounded. These models benefit from indirect causal-informational relations with worldly entities and properties, mediated by human interactions with the world that shape the linguistic inputs they receive during training and inference. Preference tuning introduces an additional learning objective that is directly world-involving, as it encompasses a requirement for truthfulness or factuality. The normativity of this requirement supplies an extra-linguistic accuracy standard that selects and stabilises the internal states that contribute to meeting it. More tentatively, we suggested that even without preference tuning, the representations and outputs of LLMs may achieve referential grounding in limited domains through pre-training and in-context learning on specific input sequences, such as those describing game moves.

This discussion led to a counter-intuitive conclusion: some multimodal models trained on images in addition to text, like CLIP, are poor candidates for referential grounding, since they lack the kind of learning process that could supply world-involving functions. Even embodied systems that combine a `frozen' LLM with a robotic body, like SayCan, fail to acquire internal states and outputs that are referentially grounded. Our analysis highlights that multimodality and embodiment are neither necessary nor sufficient conditions for referential grounding in artificial systems, although they are certainly compatible with it given the right kind of learning process.

\newpage
\bibliography{references}

\end{document}